\documentclass[a4paper]{article}
\usepackage{INTERSPEECH2019}
\usepackage{graphicx}
\usepackage{bbm}

\title{Sequence-to-Sequence Learning via Attention Transfer\\for Incremental Speech Recognition}
\name{Sashi Novitasari$^1$, Andros Tjandra$^{1,2}$ Sakriani Sakti$^{1,2}$, and Satoshi Nakamura$^{1,2}$}
\address{
  $^1$Graduate School of Information Science, Nara Institute of Science and Technology, Japan\\
  $^2$RIKEN, Center for Advanced Intelligence Project AIP, Japan}
\email{\{sashi.novitasari.si3, andros.tjandra.ai6, ssakti, s-nakamura\}@is.naist.jp}

\begin{document}

\maketitle
\begin{abstract}
Attention-based sequence-to-sequence automatic speech recognition (ASR) requires a significant delay to recognize long utterances because the output is generated after receiving entire input sequences. Although several studies recently proposed sequence mechanisms for incremental speech recognition (ISR), using different frameworks and learning algorithms is more complicated than the standard ASR model. One main reason is because the model needs to decide the incremental steps and learn the transcription that aligns with the current short speech segment. In this work, we investigate whether it is possible to employ the original architecture of attention-based ASR for ISR tasks by treating a full-utterance ASR as the teacher model and the ISR as the student model. We design an alternative student network that, instead of using a thinner or a shallower model, keeps the original architecture of the teacher model but with shorter sequences (few encoder and decoder states). Using attention transfer, the student network learns to mimic the same alignment between the current input short speech segments and the transcription. Our experiments show that by delaying the starting time of recognition process with about 1.7 sec, we can achieve comparable performance to one that needs to wait until the end.

\end{abstract}

\section{Introduction}
\vspace{-0.1cm}
The demand for speech translation systems at meetings and lectures continues to increase. Since the length of complete sentences in such talks can be long and complicated, simultaneous speech translation is required to mimic human interpreters and translate the incoming speech stream from a source language to target language in real time. One challenge for achieving simultaneous speech translation is the development of incremental ASR.

Researchers have been working on speech recognition technology for decades. A number of techniques of real-time ASR exist, especially
in the context of statistical ASR with a hidden Markov model (HMM) \cite{bamberg1990realtimeASR,hugginsdaines2006realtimeASR,platek2014onlineASR}. However, many current state-of-the-art ASR systems rely on attention-based sequence-to-sequence deep learning frameworks \cite{kim2017ctcAtt,chiu2018stateArtASR}. Today's attentional mechanisms are based on a global attention property that requires the computation of a weighted summarization of the entire input sequence generated by the encoder states. This means that the system can only generate text output after receiving the entire input speech sequence. Consequently, utilizing it in situations that require immediate recognition is difficult.

Several studies proposed local attention mechanisms \cite{bahdanau2016attASR,tjandra2017localAtt} that limit the area explored by the attention by largely reducing the total training complexity without reducing the latency. For work that enables incremental recognition of speech, Hwang and Sung employed a unidirectional RNN with a CTC acoustic model and a unidirectional RNN language model \cite{hwang2016isr}. To avoid continuous output revision, they also proposed depth-pruning in the beam-search during the output generation. Jaitly et al. proposed a neural transducer framework \cite{jaitly2016onlineSeq} that incrementally recognizes the input speech waveforms. The formulation required inferring alignments during training, and they utilized a dynamic programming algorithm to compute ``approximate" best alignments in each speech segments. Their model is strongly related to a sequence transducer that used connectionist temporal classification (CTC) \cite{graves2012rnnTransduction,graves2013isr}. The improved version of a neural transducer, which has also been discussed \cite{sainath2018improveNt}, allows the attention mechanism to look back at many previous chunks without introducing additional latency.

However, most ISR models utilize different frameworks and learning algorithms that are more complicated than the standard ASR model. One main reason is because such models need to decide incremental steps and learn the transcription that is aligned with the current short speech segment. In this work, we propose attention-transfer ISR (AT-ISR) by the following:
\begin{enumerate}
\item employing the original architecture of an attention-based ASR for ISR tasks but with shorter sequences by treating the full-utterance ASR as the teacher model and the ISR as the student model.
\item utilizing attention transfer so that the student ISR model mimics the speech-text alignment produced by the standard ASR model.
\item investigating the impact of the input size to achieve the shortest speech chunks that can still produce reliable text output.
\end{enumerate}

\begin{figure*}[ht]
  \centering
  \includegraphics[width=0.9\textwidth]{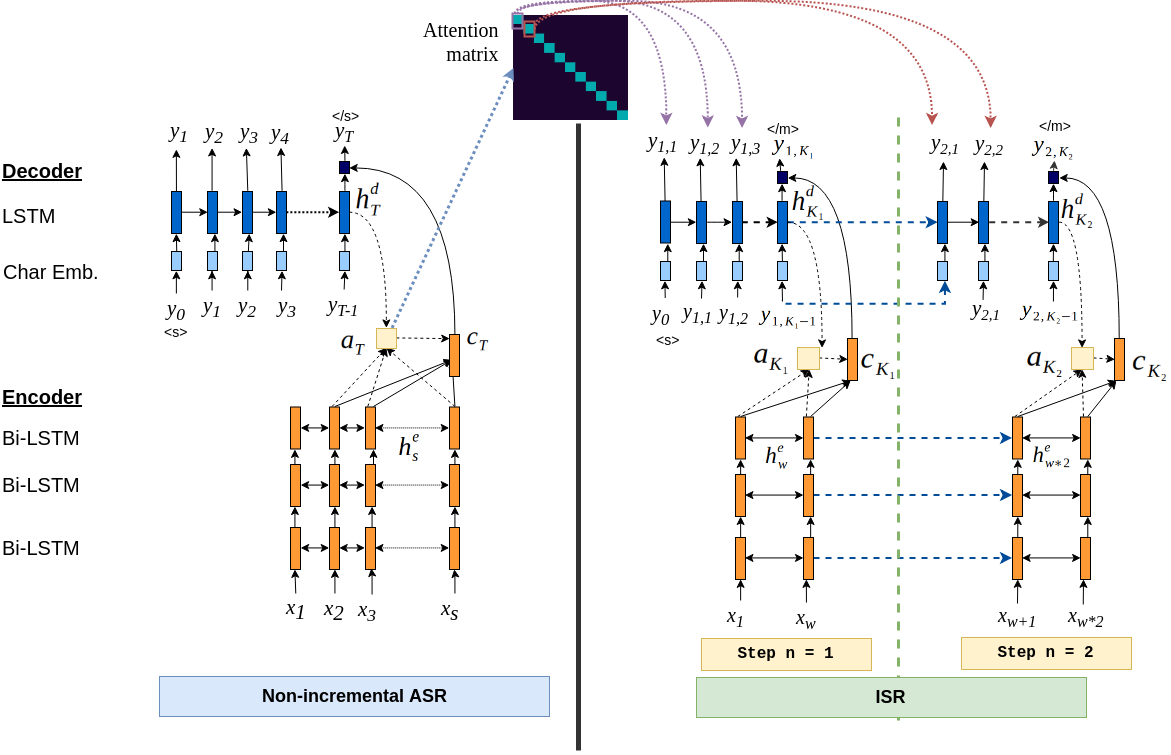}
  \caption{Training method of proposed AT-ISR: Each recognition step in ISR takes an input segment. For each step, ISR refers the attention alignment from non-incremental ASR model. If current true output is aligned to the current segment, ISR performs a decoding step for it. If previous true output is the last character that should be aligned to the current segment, ISR decodes for the end-of-block \textit{$<$/m$>$} 
   and moves to recognize the next segment. }
  \label{fig:isr}
\end{figure*}

\section{Related Works}
\vspace{-0.1cm}
A knowledge distillation method trains a student model, which is a simplification of a more complex model that acts as a teacher \cite{bucila2006compress,hinton2015distiling}. A student network is commonly constructed as a compression version that is shallower or thinner that then trains the network to mimic the original teacher network by minimizing the loss (typically L2-norm or cross-entropy) between the student and teacher output.

Another approach is attention transfer, which was recently proposed by Zagoruyko and Komodakis \cite{zagoruyko2017attTransfer} for image processing. Its basic idea is to ensure the spatial distribution of the student and teacher activations that are similar at selected layers in the network. Each layer in the student network is trained to focus on the same things as in the teacher network. Various tasks have also utilized attention transfer \cite{li2017attTransfer,yu2018attTransfer}, but not yet in ASR.

In this work, we applied attention transfer in speech recognition task by treating the full-utterance ASR as the teacher model and the ISR as the student model. Instead of using a thinner or shallower model, however, we design an alternative student network that retains the original architecture of the teacher model but with shorter sequences (only a few encoder and decoder states). In this way, no new redesign is needed for the ISR, and some hyperparameters can be used without changing them. With attention transfer, the student network learns to mimic the same alignment between the current input short speech segments and the transcription.

\section{Overview of Sequence-to-Sequence ASR Architecture}
\vspace{-0.1cm}
Our works are based on the standard non-incremental character-level sequence-to-sequence ASR \cite{chan2016las,chorowski2015attASR}. It consists of encoder, decoder, and attention modules that can directly model the conditional probability of $P({\mathbf{Y}}|{\mathbf{X}})$, where $\mathbf{X}$ is a sequence of the framed speech features with length $S$ and $\mathbf{Y}$ is a sequence of the characters with length $T$.

In this network, the encoder transforms the input speech sequence $\mathbf{X}$ to hidden representative information ${\mathbf{h}^e} = [h^e_1, ...,h^e_S]$ that will later be processed by the decoder. If downsampling is applied in the encoder, each hidden state represents several number of input frames.

The decoder attempts to predict the target sequence probability $p_{y_t}$, given the previous output $\mathbf{Y}_{<t}$, the current context information $c_t$ and the current decoder hidden state $h_t^d$. The context information $c_t$ is produced by attention modules \cite{bahdanau2014neuralAtt} at time \textit{t} based on encoder and decoder hidden states with following formula:
\begin{equation}
  c_{t} = \sum\limits_{s=1}^S a_{t}(s) * h_{s}^{e}
  \label{eq1}
\end{equation}
\begin{equation}
  a_{t}(s) = \frac{exp(Score(h_{s}^{e},h_{t}^{d}))}{\sum\limits_{s=1}^S exp(Score(h_{s}^{e},h_{t}^{d}))}
  \label{eq1}
\end{equation}

The scoring for the context can be done using one of the following scoring functions \cite{luong2015effective}:
\begin{align}
  \text{Score}(h_s^e, h_t^d) =
  \begin{cases}
  \langle h_s^e, h_t^d\rangle, & \text{dot product}  \\
  h_s^{e\intercal} W_{s} h_t^d, & \text{bilinear}  \\
  V_s^{\intercal} \tanh(W_{s} [h_s^e, h_t^d]) & \text{MLP}, \label{eq:mlpscore}  \\
  \end{cases}
  \end{align}
and $\text{Score}$ is $(\mathbb{R}^M \times \mathbb{R}^N) \rightarrow \mathbb{R}$, where $M$ is the number of encoder hidden units and $N$ is the number of decoder hidden units. The model loss function is formulated as:
\begin{equation}
\L_{ASR}(\mathbf{Y}, \mathbf{p}_Y) = -\frac{1}{T}\sum_{t=1}^{T}\sum_{c=1}^{C}\mathbbm{1}(y_t=c)\log{p_{y_t}}[c], 
\label{eq:loss_asr}
\end{equation} 
where $C$ is the number of output classes.

\section{Proposed Sequence-to-Sequence ISR}
\vspace{-0.1cm}
The proposed AT-ISR retains the original ASR architecture and performs several recognition steps incrementally with shorter sequences as shown in Fig.~\ref{fig:isr}.
In the character-based full-utterance ASR, a sequence of speech frames $\mathbf{X}$ with length $S$ transcribed as a character sequence $\mathbf{Y}=[y_1,y_2,...,y_T]$ with length $T$.
In the AT-ISR, for each recognition step, the model takes a segment of $\mathbf{X}$ with length $W$ where ($W<S$) and 
outputs an aligned segment of $\mathbf{Y}=[y_{n,1},y_{n,2},...,y_{n,K_n}]$ with maximum length $K$ where ($K_n \leq K<T$) and $n$ is the step index.
Here, the window size $W$ is equal for all steps.
%The segment recognition done sequentially until it reaches the last segment.

In the AT-ISR training, we treat the full-utterance ASR as the teacher model and the ISR as the student model.
With attention transfer, the student network will learn to mimic the same alignment between the input segments and the transcription as the teacher network.
Therefore, inferring alignments during training is not necessary anymore.
% by learning the attention-based alignments that generated by teacher.
%Therefore, inferring alignments during training is not necessary anymore.

During the training, for each incremental step $n$, AT-ISR learns to decode the characters that scored the highest monotonic alignment score to the current input segment or blocks from the distilled teacher's attention matrix. 
Attention matrix that generated by teacher consists of alignment scores of each character to each frame block. 
%With downsampling applied in the encoder side, each frame block will be consists of frames with length equals to the encoder's downsampling rate $D$. 
%This will be the alignment unit for the AT-ISR training.
Therefore, the AT-ISR input segment size $W$ will be equals to 1 frame block or multiple of it. 
%During the training, for each incremental step, AT-ISR learns to decode the characters that scored the highest alignment score to the current input block from the distilled attention matrix.

In training, if the decoding in AT-ISR reaches the last character that aligned to the current input segment, AT-ISR will learn to output an end-of-block symbol \textit{$<$/m$>$} then move to the next segment. 
In the actual recognition with AT-ISR, the decoding in each step $n$ will be done until the \textit{$<$/m$>$} symbol is predicted or the output length in step $n$ reaches $K$.
%or the output length of the current step reach the maximum length $K$.
%Here, each character only aligned to $W$ frames.
%As the actual alignment length might not equals to $W$ frames, our experiment shows the effect of the number of $\mathbf{W}$ in AT-ISR.

To accommodate the connection between each recognition step, we slightly modify the encoder and decoder parts:
\begin{itemize}
\item Encoder\\
The modification in the encoder is done on the input side. 
AT-ISR encodes $W$ frames in each step and the attention transfer during training will force the outputs to be aligned into $W$ frames.
As a character might actually be aligned to frames longer than $W$, we investigated two input settings of encoder: allow the model to look-back or look-ahead the main input block and not.
If these additional context frames are included, the outputs of a step are the characters that aligned to the main input block.

\item Decoder\\
We used \textit{$<$s$>$} and \textit{$<$/s$>$} symbols to define a sentence's beginning and end. 
In incremental recognition, we used \textit{$<$/m$>$} symbols in the transcription to define the end of the block. 
%If symbol \textit{$<$\textbackslash m$>$} is predicted, this indicates that the AT-ISR can move to recognize the next block.
In the decoding of the first input segment, AT-ISR takes \textit{$<$s$>$} symbol as the decoder's initial input.
For the decoding of subsequent segments, we experimented on two types of initial input: 
beginning-of-block \textit{$<$m$>$} and the last output from the previous incremental step. 
Similar to the mechanism in the neural transducer, we allowed the model to produce no output from the current input segment.
\end{itemize}

\section{Experimental Set-up}
\vspace{-0.1cm}
\subsection{Model Configuration}
The same attention-based encoder-decoder architecture is used for the non-incremental ASR and AT-ISR systems. 
The encoder part consists of an FNN layer and three bidirectional LSTM layers with a downsampling rate of two states in each of the LSTM layers. 
The first layer in the encoder takes a sequence of framed speech 
%with 80 features 
and outputs 512 features, and each LSTM layer outputs 256 features. 
Further details of our implementation of attention-based encoder-decoder architecture can be found here \cite{tjandra2017spchain}.

The encoder resulted in downsampled states with downsampling rate of eight. 
Therefore, in this experiment, a basic block of speech consists of eight frames (137.5 ms). 
The attention matrix from the non-incremental ASR aligned each character to each encoder state that represents eight frames. 
%In the AT-ISR system, we simply use a block of eight frames as our basic incremental step.

The decoder side consists of an embedding layer, an LSTM layer with an attention mechanism, and a softmax layer. 
Here, we also applied an attention mechanism with an MLP-scoring function that utilized previously proposed multi-scale alignment and contextual history \cite{tjandra2018multiAtt}. 
This mechanism maintains the history of the location and the contextual information of the previous time-steps, which improved the scoring function more than the standard attention. 
The standard attention applies the multiscale convolution of the previous attention vector to the current decoder state.

\subsection{Dataset and Features}
Due to resource constraints, we first used the LJ Speech corpus \cite{ito2017ljspeech} to find the best configuration of our ISR model. 
Then we applied the well-performed configurations on the Wall Street Journal (WSJ) corpus \cite{paul1992wsj}.

The LJ Speech dataset consists of 13.100 English speech utterances spoken by a single speaker (24 hours of speech). 
We simply divided the LJ Speech data into 12.314, 393, and 393 utterances as training, development, and test sets.
The complete data of the WSJ corpus are contained in an SI284 (SI84+SI200) dataset. 
We followed the Kaldi \cite{povey2011kaldi} recipe for the training (\textit{si84} and \textit{si284}), development (\textit{dev93}), and test (\textit{eval92}) sets. 
SI84 consists of 7138 utterances (16 hours of speech) spoken by 83 speakers and SI200 consists of 30,180 utterances (66 hours of speech) spoken by 200 speakers without any overlap with the speakers of SI84. 
%``dev93" denotes the development and ``eval92" denotes the test sets.

All of the utterances in both datasets have 16-kHz sampling rates. 
From the speech utterances, we extracted the 80-Mel spectrogram for each frame with a window length of 50 ms that shifted by 12.5 ms from the previous frame.

\section{Experiment Results}
\vspace{-0.1cm}
Our topline is the standard non-incremental ASR that conducted a greedy search for the decoding. 
We used an ISR without attention transfer learning as our baseline.
The baseline ISR was trained identically as the non-incremental ASR and then tested incrementally by adding a zero vector at the end of the input segment without input overlap. 
%Although the baseline ISR was trained identically as the non-incremental ASR, it was tested incrementally by adding a zero vector at the end of the input block without input overlap. 
Here, the decoding of each segment finished when the \textit{$<$/s$>$} or $<$\textit{blank}$>$ symbol (included in the vocabulary) predicted.
\begin{table}[h]
\caption{CER (\%) comparisons among topline ASR, baseline ISR, and AT-ISR with different approaches on LJ Speech}
\vspace{-0.4cm}
\label{tbl:res1}
\begin{center}
  %\resizebox{\columnwidth}{!}
  \begin{tabular}{ |l|l|c|c|c|c| }
  \hline
  \textbf{Enc-Inp} &\textbf{Dec-Inp} & \textbf{Delay (sec)} & \textbf{Dev.} & \textbf{Test}\\
  \hline
  \hline
  %\multicolumn{4}{|c|}{\textbf{Non-incremental ASR}} \\
  \hline
  \multicolumn{2}{|l|}{{Topline ASR}}& 6.54 (avg)  & 2.84 & 2.78 \\
  \multicolumn{2}{|l|}{{Baseline ISR}}& 0.14 & 79.63 &  80.34 \\
  \hline
  \hline
  \multicolumn{5}{|c|}{\textbf{AT-ISR - reset state}} \\
  \hline
  No overlap & \textit{$<$m$>$}& 0.14 & 32.51 & 32.35 \\
  No overlap & last prev. char & 0.14 & 26.15 & 26.52 \\
  Overlap & \textit{$<$m$>$} & 0.24& 23.74 & 23.40 \\
  Overlap & last prev. char & 0.24& 23.40 & 14.22 \\
  \hline
  \multicolumn{5}{|c|}{\textbf{AT-ISR - keep state}} \\
  \hline
  No overlap & \textit{$<$m$>$} & 0.14  & 24.35& 24.44 \\
  No overlap & last prev. char & 0.14 & 22.69 & 23.04 \\
  Overlap & \textit{$<$m$>$} & 0.24 & 8.83  & 8.16 \\
  Overlap & last prev. char  & 0.24 &\textbf{8.82} & \textbf{8.39} \\
  \hline
  
  \end{tabular}%}
\end{center}
\end{table}

First, we compared the performances of topline ASR, baseline ISR, and AT-ISR as shown in Table~\ref{tbl:res1}. 
Here we also investigated the performance of the AT-ISRs with or without overlapping input as well as the different symbols for decoding initial input.
% (the \textit{$<$m$>$} symbol or the output character from previous steps). 
One input segment in all ISRs in Table~\ref{tbl:res1} consist of one main input block.
%As described before, one block consists of eight frames (135.7 ms). 
In the models that allow input overlap, the input segment also includes one block ahead of the main block, resulting in 16 input frames. 
All of the models were evaluated based on the character error rate (CER).

The results in Table~\ref{tbl:res1} show that the AT-ISR models significantly outperformed the baseline and achieved CER below 10\%, similar to the topline.
Among the proposed models, keeping the model states and transferring the learning greatly improved the performance. This configuration then applied in the further experiments.
The lowest CER was achieved by allowing the model to overlap the input and by feeding the last character from the previous step as the decoder's initial input.

\begin{table}[t]
\caption{AT-ISRs CER (\%) with different number of additional blocks in each step on LJ Speech (1 main block=137.5ms)}
\vspace{-0.4cm}
\label{tbl:res2}
\begin{center}
  \begin{tabular}{ |c|c|c|c|c| }
  \hline
  \textbf{Look-back} & \textbf{Look-ahead}  & \textbf{Delay (sec)} & \textbf{Dev.} & \textbf{Test}\\
  \hline
  \hline
  \multicolumn{3}{|l|}{{Non-incremental ASR}} & 2.84 & 2.78 \\
  \hline
  \hline
  \multicolumn{5}{|c|}{{\textbf{ISR}}} \\
  \hline
   0 block & 0 block & 0.14 & 22.69  & 23.04  \\
   \hline
   0 block & 1 block & 0.24 & 8.82  & 8.39  \\
   0 block & 2 blocks & 0.34 & 5.68  &  5.32  \\
   0 block & 4 blocks &  0.54 & \textbf{4.45} & \textbf{4.33}  \\
  \hline
   1 block & 1 block & 0.34 & 7.60  & 6.86 \\
   2 blocks & 1 block & 0.44 & 7.03   & 6.80  \\
   4 blocks & 1 block & 0.64 & 7.09  &  6.89 \\
  \hline
  
  \end{tabular}
\end{center}
\end{table}
\begin{figure}[t]
  \centering
  \includegraphics[width=\linewidth]{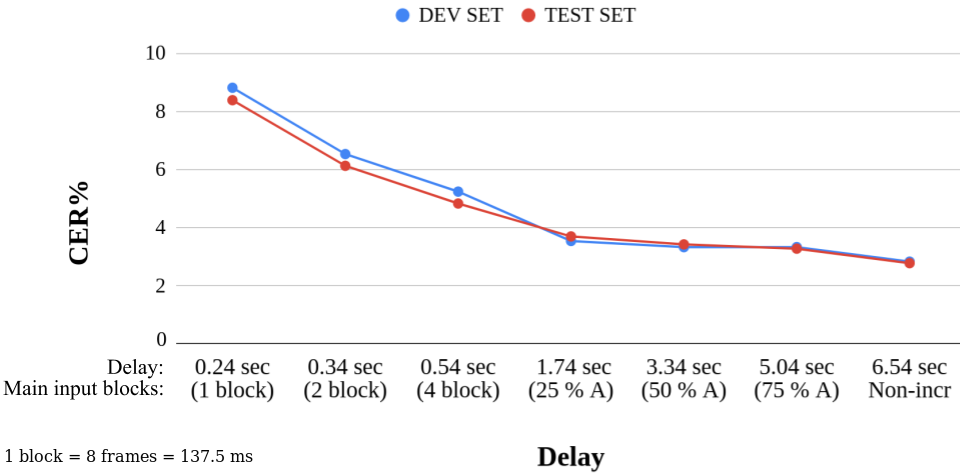}
  \caption{AT-ISRs performance on LJ Speech with different main block size, 0 look-back, and 1 look-ahead block in each step. (A = average block number of an utterance in LJ Speech)}
  \label{fig:res}
\end{figure}

We further investigated the impact of the size of the additional blocks in the AT-ISR.
% with the last character from the previous step as the decoder's initial input. 
Our results in Table~\ref{tbl:res2} reveal that looking ahead from the main block resulted in a better performance than looking back.
The AT-ISR maintains the information from the previous steps, thus adding the previous block to the main block is not necessary. 
%The ahead blocks provide new information that supports a better understanding of the main block. 
%Attention transfer forces the AT-ISR to align an output into $W$ frames, even though it actually might be aligned to longer frames.
%In such case, without ahead block, the model will lack of information and easier to make mistakes.
On the contrary, the ahead blocks able to support a better understanding of the main block by providing new information.
%Therefore, the look-ahead blocks complete the necessesary information
%for the examples are when a character actually aligned to longer frames than the main block
% or when the decision on a character depends on the further context.
By looking four blocks ahead, the AT-ISR achieved a performance with a smallest difference to the non-incremental ASR.

We also investigated the impact of the size of the main block in each recognition steps. 
Fig.~\ref{fig:res} illustrates the performances of the proposed model where a single recognition step accommodates several blocks.
% where a block consists of eight frames. 
In this figure, each model allows looking ahead to another block. 
The results show a clearer trade-off between time and performance. 
Although significant improvement happened until delay of 1.7 seconds, it did not happen similarly on subsequent sizes. 
Here, we investigated on how to reduce the delay without causing a significant decrease in ASR performance. 
Our experiments show that by delaying the starting time of recognition with about 1.7 sec, we can achieve a comparable performance to one that needs to wait until the end. 
This indicates that the student model performance with attention transfer approached that of the teacher model.

From our experiments on LJ Speech, we learned that the optimum performance in reasonable latency achieved the following: 
(1) included ahead blocks, 
(2) set the last character of the previous step as the decoder input, 
(3) kept the recurrent states across the steps, and 
(4) utilized the distilled knowledge of the attention matrix in the training. 
With this configuration, we trained the ISR models with the WSJ dataset with 1 block as the main block for short latency.
The results on  \textit{eval92} can be seen in Table~\ref{tbl:all}. 
Here, we also compared our results with several published models of non-incremental ASR.
% such as CTC, Attention Encoder-Decoder and Joint CTC-Attention model. 
Our results demonstrate that the AT-ISR can still performed with comparable performance with other published models.
In order to balance the performance and latency, our experiments show that AT-ISR delay 0.54 sec or input segment with one main input block and four look-ahead blocks in each recognition step are sufficient.

\begin{table}[h]
	\centering
	\small
	\caption{CER (\%) on eval92 set from topline ASR and AT-ISR models trained on WSJ-SI84 and WSJ-SI284. }
	\vspace{-0.3cm}
	\label{tbl:all}
	\begin{tabular}{|l|l|l|l|}
		\hline
		\multicolumn{2}{|c|}{\textbf{Models}}
		& \multicolumn{2}{c|}{\textbf{WSJ CER (\%)}} \\ \hline 
		\multicolumn{2}{|c|}{\textbf{Non-Incremental ASR (Topline)}}
        & \multicolumn{1}{|c|}{\textbf{SI84}}
        & \multicolumn{1}{|c|}{\textbf{SI284}} \\ \hline
		\multicolumn{2}{|l|} {CTC \cite{kim2017ctcAtt}}     & 20.34 & 8.97                                       \\ \hline
		\multicolumn{2}{|l|} {Att Enc-Dec Content \cite{kim2017ctcAtt}}  & 20.06 & 11.08                                       \\ \hline
		\multicolumn{2}{|l|} {Att Enc-Dec Location \cite{kim2017ctcAtt}}  & 17.01 & 8.17                                       \\ \hline
		\multicolumn{2}{|l|} {Joint CTC+Att (MTL) \cite{kim2017ctcAtt}} & 14.53 & 7.36                 \\ \hline
		\multicolumn{2}{|l|} {Att Enc-Dec (ours)} & 17.05 & 6.80                                             \\ \hline \hline
		\multicolumn{2}{|c|}{\textbf{AT-ISR}} & & \\
        \cline{1-2}
        \multicolumn{1}{|c|}{\textbf{Look-back}} & \multicolumn{1}{|c|}{\textbf{Look-ahead}}
        & \multicolumn{1}{|c|}{\textbf{SI84}}
        & \multicolumn{1}{|c|}{\textbf{SI284}} \\ \hline
        0 block & 1 block & 30.81 & 19.78 \\
        0 block & 4 blocks &  \textbf{18.05} & \textbf{9.06} \\
        \hline 
          \multicolumn{4}{|l|}{*Same delay configuration as Table 2} \\
\hline
	\end{tabular}
\end{table}

\vspace{-0.3cm}
\section{Conclusions}
\vspace{-0.1cm}
We constructed a character-level AT-ISR framework that was trained with the original architecture of the attention-based sequence-to-sequence ASR model. 
The main difference is that it consists of shorter sequences than the standard architecture. 
No new redesign was needed for the ISR, and some hyperparameters can be used without any changes.  
Transfer learning treats the non-incremental ASR model as the teacher and the ISR as the student model. 
Student ISR learns the same attention alignment as the teacher model's, allowing a simple mechanism in the incremental recognition. 
%both of which utilize the same attention alignment with the teacher model's, allowing a simple mechanism in the incremental recognition. 
Various types of models have been explored. 
The optimum performance was achieved by including a few ahead blocks, setting the last character of the last set as the decoder input, keeping the recurrent states across the steps, and utilizing the attention transfer.

\section{Acknowledgements}
Part of this work was supported by JSPS KAKENHI Grant Numbers JP17H06101 and JP17K00237.

\bibliographystyle{IEEEtran}
\bibliography{mybib}

\end{document}